\pdfoutput=1

\documentclass[a4paper,conference]{IEEEtran}
\usepackage{kpi-relation-extraction}
\ifCLASSINFOpdf
\else
\fi
\makeatletter
\def\ps@IEEEtitlepagestyle{%
  \def\@oddfoot{\mycopyrightnotice}%
}
\def\mycopyrightnotice{%
\fbox{\parbox{\dimexpr\textwidth-2\fboxsep-2\fboxrule\relax}{
\begin{minipage}{\textwidth-2\fboxsep-2\fboxrule}
  \footnotesize
  \textcopyright 2022 IEEE. Personal use of this material is permitted. Permission from IEEE must be obtained for all other uses, in any current or future media, including reprinting/republishing this material for advertising or promotional purposes, creating new collective works, for resale or redistribution to servers or lists, or reuse of any copyrighted component of this work in other works.
  \end{minipage}
}}
}
\makeatother

\begin{document}
%
\title{KPI-BERT: A Joint Named Entity Recognition and Relation Extraction Model for Financial Reports}



\author{\IEEEauthorblockN{Lars  Hillebrand\IEEEauthorrefmark{1}\IEEEauthorrefmark{2}\IEEEauthorrefmark{3}, Tobias Deußer\IEEEauthorrefmark{2}\IEEEauthorrefmark{3},
Tim Dilmaghani\IEEEauthorrefmark{4}, Bernd Kliem\IEEEauthorrefmark{4}, \\ Rüdiger Loitz\IEEEauthorrefmark{4}, Christian Bauckhage\IEEEauthorrefmark{2}\IEEEauthorrefmark{3}, Rafet Sifa\IEEEauthorrefmark{2}}
\IEEEauthorblockA{\IEEEauthorrefmark{2}\textit{Fraunhofer IAIS}, Bonn, Germany \\
\IEEEauthorrefmark{3}\textit{University of Bonn}, Bonn, Germany\\
\IEEEauthorrefmark{4}\textit{PricewaterhouseCoopers GmbH}, Düsseldorf, Germany}
}

\maketitle


\newenvironment{symbolfootnotes}
  {\par\edef\savedfootnotenumber{\number\value{footnote}}
  \renewcommand{\thefootnote}{\fnsymbol{footnote}}
  \setcounter{footnote}{0}}
  {\par\setcounter{footnote}{0}}
\begin{symbolfootnotes}
\footnotetext[1]{Correspondence to
\href{mailto:lars.patrick.hillebrand@iais.fraunhofer.de}{lars.patrick.hillebrand@iais.fraunhofer.de}. \\ ORCID ID: \href{https://orcid.org/0000-0002-5496-4177}{0000-0002-5496-4177}.}
\end{symbolfootnotes}

\begin{abstract}
We present KPI-BERT, a system which employs novel methods of named entity recognition (NER) and relation extraction (RE) to extract and link key performance indicators (KPIs), e.g. ``revenue'' or ``interest expenses'', of companies from real-world German financial documents. 
Specifically, we introduce an end-to-end trainable architecture that is based on Bidirectional Encoder Representations from Transformers (BERT) combining a recurrent neural network (RNN) with conditional label masking to sequentially tag entities before it classifies their relations. Our model also introduces a learnable RNN-based pooling mechanism and incorporates domain expert knowledge by explicitly filtering impossible relations.
We achieve a substantially higher prediction performance on a new practical dataset of German financial reports, outperforming several strong baselines including a competing state-of-the-art span-based entity tagging approach.
\end{abstract}


%
\IEEEpeerreviewmaketitle

\section{Introduction}
In the context of business administration, key performance indicators (KPIs) are defined as quantitative measures about structural entities and are usually utilized for facilitating descriptive, comparative and predictive analysis as well as for informed decision making \cite{krause2009controlling,Brito2019AHA}.
Considering the latter, (semi-)automatically extracting information (e.g. in form of values or relationships) related to such indicators can give companies competitive advantages due to the time efficiency practitioners gain, especially when analyzing large amount of data. Recently Natural Language Processing and Machine Learning based approaches have been deployed to extract KPI-related information from unstructured data, such as financial documents. These approaches have also been used to support financial auditors with certain elementary processes related to analyzing and comparing information from single as well as multiple documents \cite{sifa2019towards}.
Although being successfully deployed, these concepts often suffer from either being rule-based and inflexible \cite{farmakiotou2000rule}, only considering structured data (i.e. tables) \cite{Brito2019AHA}, or focusing exclusively on numerical cross checking \cite{Cao2018TowardsAN}.

To alleviate these challenges, we present KPI-BERT, an automated system which leverages new methods of named entity recognition (NER) and relation extraction (RE) to detect KPIs and their relationships in real-world German financial documents. The described system is currently being integrated in the auditing process of a major auditing company and promises to achieve significant efficiency gains. 

Given the following sentence from a financial statement,
\begin{tcolorbox}[breakable,notitle,boxrule=0pt,
boxsep=0pt,left=0.6em,right=0.6em,top=0.5em,bottom=0.5em,
colback=gray!10,
colframe=gray!10]
``In 2021 the $\underset{\text{{\color{ForestGreen}{kpi}}}}{\text{{\color{ForestGreen}{revenue}}}}$ increased to \$$\underset{\text{{\color{ForestGreen}{cy}}}}{\text{{\color{ForestGreen}{100}}}}$ million (prior year: \$$\underset{\text{{\color{ForestGreen}{py}}}}{\text{{\color{ForestGreen}{80}}}}$ million) while the $\underset{\text{{\color{ForestGreen}{kpi}}}}{\text{{\color{ForestGreen}{total costs}}}}$ decreased to \$$\underset{\text{{\color{ForestGreen}{cy}}}}{\text{{\color{ForestGreen}{50}}}}$ million  (prior year: \$$\underset{\text{{\color{ForestGreen}{py}}}}{\text{{\color{ForestGreen}{70}}}}$ million).''
\end{tcolorbox}
it automatically recognizes and classifies the highlighted named entities and links their relations: 
\begin{tcolorbox}[breakable,notitle,boxrule=0pt,
boxsep=0pt,left=0.6em,right=0.6em,top=0.5em,bottom=0.5em,
colback=gray!10,
colframe=gray!10]
\small
$\underset{\text{{\color{ForestGreen}{kpi}}}}{\text{{\color{ForestGreen}{revenue}}}} - \underset{\text{{\color{ForestGreen}{cy}}}}{\text{{\color{ForestGreen}{100}}}}$, $\underset{\text{{\color{ForestGreen}{kpi}}}}{\text{{\color{ForestGreen}{revenue}}}} - \underset{\text{{\color{ForestGreen}{py}}}}{\text{{\color{ForestGreen}{80}}}}$, $\underset{\text{{\color{ForestGreen}{kpi}}}}{\text{{\color{ForestGreen}{total costs}}}} - \underset{\text{{\color{ForestGreen}{cy}}}}{\text{{\color{ForestGreen}{50}}}}$,
$\underset{\text{{\color{ForestGreen}{kpi}}}}{\text{{\color{ForestGreen}{total costs}}}} - \underset{\text{{\color{ForestGreen}{py}}}}{\text{{\color{ForestGreen}{70}}}}$
\end{tcolorbox}
where \textit{kpi}, \textit{cy} (current year value) and \textit{py} (prior year value) are defined entity classes explained in Table \ref{tab:entities}.
In particular, the system utilizes a BERT-based \cite{devlin2018bert} architecture that novelly combines a recurrent neural network (RNN) with conditional label masking to sequentially tag the above emphasized entities before it classifies the linked relations. We further improve the setup by employing trainable RNN-based pooling layers, which outperform the established mean- and max-pooling counterparts. The model also incorporates domain expert knowledge into the process. First, it filters impossible relation candidates prior to their classification since not all entity pair combinations are allowed to be linked (see Table \ref{tab:relations}). Second, we post-process the predicted relations by removing overlapping ones based on their prediction probability. 

We benchmark our approach against multiple strong baselines, which also build on BERT-encoded word embeddings but utilize different entity tagging schemes, namely state-of-the-art span-based tagging \cite{eberts2019span}, sequential Conditional Random Field (CRF) tagging \cite{huang2015bidirectional} and standard linear tagging \cite{taille2020let}. In addition, we thoroughly investigate the impact of various parameter ablations, including the usage of different word pooling functions. 
We find that our system outperforms the competing architectures in robustly extracting and relating KPIs within financial reports.



In summary, our contributions are twofold:
\begin{itemize}
    \item We present a novel system that automatically extracts and links key performance indicators (KPIs) from financial documents and is actively integrated in the auditing process of a major auditing firm.
    \item We introduce a new BERT-based architecture that employs a gated recurrent unit (GRU) coupled with trainable pooling layers and conditional label masking to successfully address the sequential nature of the KPI extraction task.
\end{itemize}

In the following, we first review related work and recent advances in named entity recognition and relation extraction. Next, Section~\ref{sec:method} introduces our model, competing baselines and the corresponding training process. In Section~\ref{sec:experiments}, we describe our real-world dataset of financial documents and present the experimental setup along with performance results of all models. We close with concluding remarks and an outlook into conceivable future work.

\section{Related Work}
\label{sec:related}

In this work, we focus on our specific setup of token-level entity tagging combined with conditional label masking to jointly extract entities and relations on a novel corpus of German financial documents and contrast the results with various ablation studies. However, many recent studies have investigated the task of named entity recognition
(\cite{lester2020constrained}, 
\cite{Ushio2021TNERAA}, 
\cite{wang2021ImprovingNE}, 
\cite{wang2020automated})
relation extraction (\cite{Xu2021EntitySW}, \cite{Zeng2020DoubleGB}, \cite{Zhang2021DocumentlevelRE}), and the joint combination of both
(\cite{liang2020bond}, 
\cite{Shen2021ATM}, 
\cite{Wang2020TwoAB}
\cite{Ye2021PackTE}, 
\cite{Zhong2021AFE})\footnote{\cite{taille2020let} wrote a more comprehensive article on recent developments in the field of relation extraction and compared their results in depth.}.

Much effort has been spent on the task of separately recognizing named entities and extracting relations in the past, whose results were then hierarchically combined in a pipeline (\cite{fundel2007relex}, \cite{gurulingappa2012extraction}). The previously mentioned studies showed that learning these tasks jointly can improve the performance immensely and thus suggest that insightful information from one task can be exploited by the other. Furthermore, most contemporary models have their foundation in modern pre-trained natural language models 
(\cite{devlin2018bert}, 
\cite{Liu2019RoBERTaAR}, 
\cite{Radford2018ImprovingLU}, 
\cite{Radford2019LanguageMA}).
Highlighting a few of these contemporary studies, \cite{eberts2019span} introduced a model called SpERT and reported state-of-the-art results on various datasets designated for this task. \cite{giorgi2019end} leveraged BERT at its core to implement an end-to-end model on the token-level with feed forward layers for each task, achieving comparable results to \cite{eberts2019span}. \cite{liang2020bond} lessened the required annotations during the NER task by introducing a self-training approach and \cite{pappu2017lightweight} focused on diminishing the computational complexity by utilizing more compact entity embeddings.

Looking at our specific task of retrieving information from financial reports using machine learning methods, \cite{treigueiros1991application} employed a multilayer perceptron (MLP) to capture interpretable structures similar to accounting ratios.
However, the inputs for their MLP were already extracted and transformed accounting variables. A step into the direction of automatically extracting these variables has been done by \cite{farmakiotou2000rule}, who developed a NER model with a rule-based approach. The most up-to-date work in this specific field is \cite{Cao2018TowardsAN}, which leveraged a joint entity and relation extraction approach to cross-check various financial formulas. 

With respect to our domain of processing German accounting and financial documents, \cite{biesner2020leveraging} leveraged contextualized NLP methods to recognize named entities in the context of anonymization. Besides, \cite{sifa2019towards} presented a recommender-based tool that greatly simplifies and to a large extend automates the auditing of financial statements.



\section{Methodology} \label{sec:model}
\label{sec:method}



Our proposed model comprises three stacked components that we train jointly in an end-to-end fashion via gradient descent. First, a BERT-based encoder embedds the sentence into latent space. Second, a GRU-based named entity recognition (NER) decoder sequentially classifies entities using conditional label masking along with the prior tagging history. Third, a relation extraction (RE) decoder links the predicted entities.

\subsection{BERT-based Sentence Encoder}
\label{encoder}

Given a WordPiece \cite{schuster2012japanese} tokenized input sentence of $n$ subwords we use a pre-trained BERT \cite{devlin2018bert} model to obtain a sequence of $n+1$ encoded token embeddings, $(\mat{c}, \mat{t}_1, \mat{t}_2, \dots, \mat{t}_n)$, where $\mat{c} \in \RR^d$ represents the context embedding for the whole sentence and $\mat{t}_i \in \RR^d$ represents the token embedding at position $i$. To easily utilize our word-level entity annotations and to reduce model complexity, we apply a pooling function, $\text{pool}(\cdot)$, which creates word representations by combining their individual subword embeddings. Specifically, the $j$-th word, consisting of $k$ subwords, is represented as
\begin{align}
    \mat{e}_j := \text{pool}(\mat{t}_i, \mat{t}_{i+1}, \dots, \mat{t}_{i+k-1}), \quad \mat{e}_j \in \RR^d.
    \label{eq:pooling}
\end{align}

While also evaluating max- and mean-pooling we employ a more sophisticated trainable RNN-pooling mechanism building on a bidirectional GRU.


In particular, the subword embedding sequence $(\mat{t}_i, \mat{t}_{i+1}, \dots, \mat{t}_{i+k-1})$ is passed bidirectionally through a forward and backward GRU, yielding the final hidden states 
\begin{align}
    \mat{h}_j^f &= \text{GRU}^f\left(\mat{t}_i, \mat{t}_{i+1}, \dots, \mat{t}_{i+k-1}\right),\\
    \mat{h}_j^b &= \text{GRU}^b\left(\mat{t}_{i+k-1}, \dots, \mat{t}_{i+1}, \mat{t}_i\right),
\end{align}
where $\mat{h}_j^f, \mat{h}_j^b \in \RR^{d/2}$ and the superscripts $\cdot^f$ and $\cdot^b$ refer to the forward and backward model, respectively. Next, we simply concatenate $\mat{h}_j^f$ and $\mat{h}_j^b$ to obtain
\begin{align}
    \mat{e}_j = \left[ \mat{h}_j^f; \mat{h}_j^b \right].
\end{align}

\subsection{NER Decoder}
\label{ner-decoder}

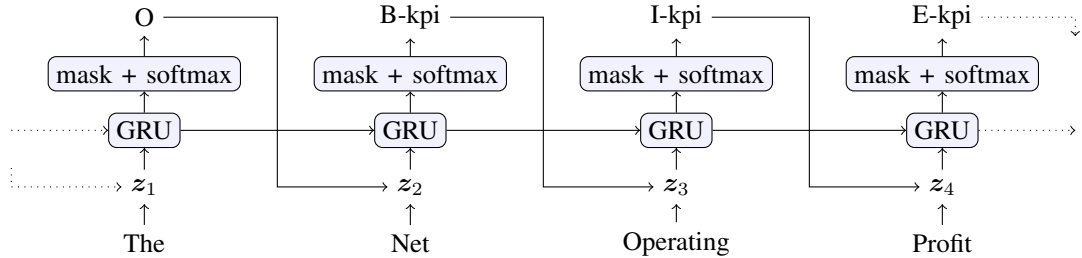
\begin{figure*}[!ht]
  \centering
   \begin{tikzpicture}[scale=1]
    
    \foreach \t [count=\step from 1] in {The,Net,Operating,Profit} {
        \node[align=center] (word\step) at (3.5*\step, -1.5) {\t};
    }
    
    \foreach \t [count=\step from 1] in {O,B-kpi,I-kpi,E-kpi} {
        \node[align=center] (label\step) at (3.5*\step, 1.5) {\t};
    }
    
    \foreach \step in {1,..., 4} {
        \node[rectangle, draw, inner sep=0.3em, fill=blue!5, rounded corners=0.3em] (GRU\step) at (3.5*\step, 0) {GRU};
        \node[rectangle, draw, inner sep=0.3em, fill=blue!5, rounded corners=0.3em] (mask\step) at (3.5*\step, 0.75) {mask + softmax};
        \node (wordembedding\step) at (3.5*\step, -0.75) {$\mat{z}_{\step}$};
        
        \draw[->] (word\step.north) -> (wordembedding\step.south);
        \draw[->] (wordembedding\step.north) -> (GRU\step.south);
        \draw[->] (GRU\step.north) -> (mask\step.south);
        \draw[->] (mask\step.north) -> (label\step.south);
    }
    
    \foreach \step in {1,...,3} {
        \pgfmathtruncatemacro{\next}{add(\step,1)}
        \draw[->] (GRU\step.east) -> (GRU\next.west);
        \draw[->] (label\step.east) -| (3.5*\step + 1.75,0) |- (wordembedding\next.west);
    }
    
    \draw[->, dotted] (1.75,-0.5) |- (wordembedding1.west);
    \draw[->, dotted] (label4.east) -| (3.5*4 + 1.75, 1.25);
    \draw[->, dotted] (1.75, 0) -- (GRU1.west);
    \draw[->, dotted] (GRU4.east) -- (3.5*4 + 1.75, 0);
    
\end{tikzpicture}
  \caption{Sequential IOBES tagging (each entity class is prepended with the prefixes  \textit{I-} (inside), \textit{B-} (begin), \textit{E-} (end) or \textit{S-} (single), while \textit{O} (outside) represents the \textit{none} class) leveraging a gated recurrent unit (GRU) and conditional label masking. $\mat{z}_j$ represents the concatenation of the previously predicted label embedding with the word embedding at position $j$. Along with the previous hidden state vector it gets passed to a GRU that, followed by label masking and a softmax layer, predicts the next IOBES tag.}
  \label{rnn-iobes-decoding}
\end{figure*}
Utilizing the BERT-encoded and pooled word embedding sequence $(\mat{e}_1, \mat{e}_2, \dots, \mat{e}_m)$, a NER decoder module classifies named entities within the sentence. 
Specifically, we introduce a sequential GRU-based tagger with conditional label masking, which builds on the IOBES annotation scheme.
IOBES tagging refers to classifying named entities on a word-level by prepending all entity classes with the prefixes \textit{I-} (inside), \textit{B-} (begin), \textit{E-} (end) and \textit{S-} (single). Additionally, \textit{O} (outside) represents the \textit{none} type in this annotation scheme. If we denote $\mathcal{E}$ as the set of possible entity types described in Table \ref{tab:entities} including the \textit{none} class, the actual number of IOBES entity tags increases to $\lvert \mathcal{E}_{\text{IOBES}} \rvert = 4 \left( \lvert \mathcal{E} \rvert - 1 \right) + 1 $. 

Given the sentence ``The Net Operating Profit increased to \$ 1.2 million in 2020 .'', IOBES tagging aims to predict the following label sequence: ``O, B-kpi, I-kpi, E-kpi, O, O, O, S-cy, O, O, O, O''.



To take the sequential nature of entity tagging into account we employ a GRU in combination with conditional label masking to sequentially predict IOBES tags considering the past predictions.
Figure \ref{rnn-iobes-decoding} visualizes this decoding scheme, which is described in more detail in the following paragraphs.

First, we define an embedding matrix $\mat{W}_{\text{label}} \in \RR^{\lvert \mathcal{E}_{\text{IOBES}} \rvert \times u}$ that holds learnable $u$-dimensional embeddings of all IOBES entity types. 

Second, we concatenate $\mat{e}_j$ with $\mat{w}_{j-1}^\text{label}$, which yields the decoding input representation of word $j$, $\mat{z}_j = \left[ \mat{e}_j; \mat{w}_{j-1}^\text{label}\right]$,
where $\mat{w}_{j-1}^\text{label} \in \RR^u$ represents the embedding of the previously predicted IOBES tag. We define $\mat{w}_0^\text{label}$ as the \textit{O} label embedding since using a dedicated begin-of-sequence embedding did not yield improved empirical results. 

Third, we feed $\mat{z}_j$ alongside the previous hidden state $\mat{h}_{j-1}$ into a GRU, yielding
\begin{align}
    \mat{h}_j = \text{GRU}\left( \mat{z}_j, \mat{h}_{j-1}\right).
\end{align}
To get IOBES tag posteriors for word $j$ we linearly transform $\mat{h}_j$ followed by masking out impossible tag predictions and applying softmax:
\begin{align}
    \mat{\hat{y}}_j = \text{softmax}\left( \text{mask}\left(\mat{W}_{\text{seq}} \mat{h}_j + \mat{b}_{\text{seq}}\right) \right).
\end{align}
Note that masking is applied conditional on the last predicted tag $\mat{\hat{y}}_{j-1}$. Specifically, if $\argmax(\mat{\hat{y}}_{j-1})$ equals \textit{O} or has prefix \textit{S} or \textit{E} we mask out all entity types with prefix \textit{I} and \textit{E}. Likewise, if $\argmax(\mat{\hat{y}}_{j-1})$ starts with \textit{B} or \textit{I} we know the next predicted tag has to be of the same entity type with prefix \textit{I} or \textit{E}. Hence, all other entity types are masked out, which effectively reduces the tagging decision to a binary classification problem.

Next, we convert all predicted IOBES tags and their word embeddings $\mat{e}_j$ to the entity-level by applying the same pooling function as in Equation \eqref{eq:pooling}. Finally, we concatenate this pooled entity representation with a span size embedding $\mat{w}_{k}^\text{width}$. It is taken from a dedicated embedding matrix $\mat{W}_{\text{width}} \in \RR^{l \times v}$ holding fixed-size embeddings of dimensionality $v$ for each span length from 1 to $l$ and is learned during training to let the model include a prior over span widths. This gives us the embedding for each entity $s$
\begin{equation}
    \mat{e}(s) := \left[\text{pool}(\mat{e}_j, \mat{e}_{j+1}, \dots, \mat{e}_{j+k-1});\mat{w}_{k}^\text{width}\right],
    \quad \mat{e}(s) \in \RR^{d+v}.
    \label{eq:span_representation}
\end{equation}

\subsection{RE Decoder}

We only allow for a single relation type between two entities, namely the \textit{matches} relation. This relation is symmetric, as it does not matter whether a KPI is matched to its value or the reverse case of a value being matched to its KPI. Additionally, we refine the entity sampling process to only allow for valid entity pairs. The \textit{relation matrix} shown in Table \ref{tab:relations} specifies which entity combinations are allowed. 
\begin{table}[t]
\scriptsize

\rowcolors{2}{gray!10}{white}
\centering
\begin{tabular}{l@{\hspace{\tabcolsep}}c@{\hspace{\tabcolsep}}c@{\hspace{\tabcolsep}}c@{\hspace{\tabcolsep}}c@{\hspace{\tabcolsep}}c@{\hspace{\tabcolsep}}c@{\hspace{\tabcolsep}}c@{\hspace{\tabcolsep}}c}
\toprule
{} &  kpi &  cy &  py & increase & decrease & davon & davon-cy & davon-py\\
\midrule
kpi            & - & 1:1 & 1:1 & 1:1 & 1:1 & 1:n & - & - \\
cy             & 1:1 & - & - & - & - & - & - & - \\
py             & 1:1 & - & - & - & - & - & - & - \\
increase       & 1:1 & - & - & - & - & - & - & - \\
decrease       & 1:1 & - & - & - & - & - & - & - \\
davon          & n:1 & - & - & - & - & - & 1:1 & 1:1 \\
davon-cy       & - & - & - & - & - & 1:1 & - & - \\
davon-py       & - & - & - & - & - & 1:1 & - & - \\
\bottomrule
\end{tabular}
\caption{Comprehensive overview of all allowed relations and their uniqueness. ``1:1'': One entity of type 1 can only be linked to one entity of type 2, ``1:n'': One entity of type 1 can be linked to many entities of type 2. ``-'': No relation possible.}
\label{tab:relations}
\end{table}
Finally, we enforce the uniqueness conditions specified in said table after the model has processed the input data. This prunes the results by eliminating relations of two entities if the same combination is predicted with a higher score in the same input sequence and such a combination is labeled as unique, i.e. a 1:1 relation. For instance, if two KPI entities are linked to a singular current year value, we only keep the relation with the higher score and discard the other.

Similar to other studies, we sample candidate pairs $s_1$ and $s_2$ from the pool $S \times S$ representing all \textit{allowed} entity combinations in the sentence. Given two entities, we concatenate their respective representations (see Equation \eqref{eq:span_representation}) with a localized context embedding $\mat{c}_{\text{loc}}$. Different from the global sentence context $\mat{c}$, $\mat{c}_{\text{loc}}$ is defined as the pooled representation\footnote{The same pooling function as in Equation \eqref{eq:pooling} is applied.} of BERT-encoded word embeddings located between $s_1$ and $s_2$. As in \cite{eberts2019span} we find that this localized context embedding is better suited for the relation classification task than the BERT context token $\mat{c}$.
Hence, we define
\begin{equation}
    \mat{x}_r(s_1, s_2) := \left[\mat{e}(s_1);\mat{c}_{\text{loc}}(s_1, s_2); \mat{e}(s_2) \right]
\end{equation}
as input for the relation classifier, where $\mat{x}_r(s_1, s_2) \in \RR^{3d + 2v}$. Due to our relation type being symmetric, we do not have to classify the inverse $\mat{x}_r(s_2, s_1)$.

The relation classifier is then defined as
\begin{equation}
    \hat{y}_r = \text{sigmoid} \left( \mat{w}_{\text{rel}}^T \mat{x}_r(s_1, s_2)  + b_{\text{rel}} \right),
    \label{eq:relation_classifier}
\end{equation}
where $\mat{w}_{\text{rel}} \in \RR^{3d + 2v}$ and $b_{\text{rel}} \in \RR$.
If the output of Equation \eqref{eq:relation_classifier} exceeds a confidence threshold $\alpha$, we consider entity $s_1$ and entity $s_2$ to match. 

\subsection{Training}


We train the above described model architecture end-to-end, including fine-tuning BERT, by minimizing the joint entity and relation classification loss defined as
$\mathcal{L} = \mathcal{L}_\text{ner} + \mathcal{L}_\text{rel}$, 
where $\mathcal{L}_\text{ner}$ denotes the categorical cross entropy loss over IOBES-tagged entity classes and $\mathcal{L}_\text{rel}$ denotes the binary cross entropy loss over the relation prediction. 

For the GRU-based NER decoder, we utilize teacher forcing to foster training convergence and stability. Thus, we embed the annotated ground truth tag and use it to condition the label masking instead of the previously predicted tag.

For the relation classifier, we utilize annotated ground truth relations as positive examples as well and randomly sample $N_\text{rel}$ negative examples from allowed ground truth entity pairs $\mathcal{S}_{\text{gt}} \times \mathcal{S}_{\text{gt}}$ that don't constitute a labeled relation.


\section{Experiments}
\label{sec:experiments}

In the following sections, we introduce our custom dataset, describe the training setup and model selection process, and evaluate results. All experiments are conducted on four Nvidia Tesla V100 GPUs and the model plus training code is implemented in PyTorch.

\subsection{Data}

Our dataset\footnote{We are currently unable to publish the dataset and the accompanying python code because both are developed and used in the context of an industrial project.} is comprised of $500$ manually annotated financial documents containing a total of $15394$ sentences and was sourced from the \textit{Bundesanzeiger}\footnote{\url{https://www.bundesanzeiger.de/}}, a platform hosted by the German department of Justice where companies publish their legally mandated documents.
\begin{table}[t]
\scriptsize
\renewcommand\tabularxcolumn[1]{m{#1}}
\rowcolors{2}{gray!10}{white}
\begin{tabularx}{\linewidth}{l@{\hspace{1em}}r@{\hspace{1em}}X} 
\toprule
\rowcolor{white} Entity &  Support & Description \\
\midrule
kpi             & $16849$ & Key Performance Indicators expressible in numerical and monetary value, e.g. revenue or net sales.\\
cy              & $11498$ & Current Year monetary value of a KPI
.
\\
py              & $5057$ & Prior Year monetary value of a KPI.\\
increase        & $356$ & Increase of a KPI from the previous year to the current year.\\
decrease        & $230$ & Decrease of a KPI from the previous year to the current year.\\
davon           & $8827$ & Davon, German for thereof, represents a \textit{subordinate} KPI, i.e. if a KPI is part of another, broader KPI.\\
davon-cy        & $8443$ & Current Year value of a thereof KPI.\\
davon-py        & $4382$ & Prior Year value of a thereof KPI.\\
\bottomrule
\end{tabularx}

\caption{Description and support of all entity types in the complete dataset, excluding the \textit{none} type.}
\label{tab:entities}
\end{table}

In a first pre-processing step, the reports are tokenized on a sentence level and subsequently on a word level using the \texttt{syntok} python library. Second, we tag monetary numbers and extract their scale (e.g. million) and unit (e.g. \$) applying rule-based string matching heuristics. Third, we filter each tokenized report for sentences containing said monetary numbers because our only interest lies in matching KPI entities with their monetary values. Next, we manually generate token and span-level annotations that are composed of an entity and a relation part, where the entity annotation signals the type of each span in a sentence and the relation annotation which entity spans are linked together. 


The manual annotations were done by a group of six qualified auditors, led by a senior auditing expert. In consultation with them, we defined the entity classes outlined in Table \ref{tab:entities}, which also shows the overall support of each class in the dataset. Thoughout the annotation process, the exact entity class definitions were refined in several iterations to adjust for variation and edge cases in the data. Most notably, we paid special attention to distinguish \textit{kpi} and \textit{davon} entities which proved difficult depending on the sentence context. After completing the annotations, the aforementioned senior auditing expert reviewed 50 randomly sampled documents and verified their quality. 
Due to budget and time constraints each document was annotated just once by a single auditor. Hence, no inter-annotator agreement metrics can be provided. Although not being entirely free of mistakes, we are confident of the overall annotation quality of the dataset.


We randomly split the pre-processed dataset on a document level into a training, validation, and test set, encompassing $13835$, $821$, and $738$ sentences each. 


\subsection{Baselines}

We compare our proposed model with three competing architectures, which all build on the BERT-based sentence encoder outlined in Section \ref{encoder}, ensuring a level playing field.

First, we replace the GRU-based NER decoder with a fully connected linear layer that classifies named entities in parallel using the BERT-encoded word embeddings as input. The resulting model was introduced by \cite{taille2020let} and functions as a straightforward baseline since it neglects inter-label dependencies when classifying entity tags.

Second, we integrate SpERT \cite{eberts2019span} in our training framework by utilizing its span-based NER decoder. Span-level entity tagging does not make use of the IOBES annotation scheme but classifies entire word spans at once. For further details we refer the interested reader to \cite{eberts2019span}. Our implementation closely follows the original code\footnote{\href{https://github.com/lavis-nlp/spert}{https://github.com/lavis-nlp/spert}.} with the exception of extending the hyperparameter search to our novel Bi-GRU pooling function and including the options to filter overlapping and impossible relations.

Third, we implement a Conditional Random Field (CRF) leveraging viterbi decoding \cite{viterbi1967error} to classify named entites, which is a popular choice for NER due to its ability of modeling label dependencies. To ensure a fair comparison with our model, we also incorporate the IOBES label constraints from Section \ref{ner-decoder} in the CRF by masking out invalid class transitions in the trainable transition matrix.




\subsection{Training Setup and Hyperparameter Selection}

To determine the best hyperparameter setup for each model we conduct an extensive grid search evaluating various parameter combinations based on the validation set relation classification F$_1$-score. A relation is considered correct if the spans and the types of both related entities are predicted correctly. Table \ref{tab:grid_search} shows all tuned model parameters with their respective ranges of values. The overall best performing setup on the validation set is highlighted in boldface. Also, note that the ``NER decoding'' row effectively discriminates KPI-BERT (GRU$_\text{LM}$ -- GRU with conditional label masking) from the other baselines.
\begin{table}[t]
\centering

\rowcolors{2}{gray!10}{white}
\scriptsize
\begin{tabular}{lc}
\toprule
Hyperparameter &  Configurations \\
\midrule
Word-, entity- and context pooling $(\text{pool})$    & \textbf{Bi-GRU}, Min, Max\\
NER decoding    & \textbf{GRU}$_{\textbf{LM}}$, CRF$_{\text{LM}}$, Span, Linear \\
Conditional label masking (LM)   & \textbf{True}, False \\
Dropout         & $0.0$, $\bm{0.1}$, $0.2$, $0.3$ \\
Confidence threshold $(\alpha)$ & $0.4$, $\bm{0.5}$, $0.6$ \\
Filtering impossible relations & \textbf{True}, False \\
Removing overlapping relations & \textbf{True}, False \\
Batch size      & $\bm{2}$, $4$, $8$ \\
Learning rate   & $5e^{-5}$, $\bm{1e^{-5}}$, $5e^{-6}$ \\
Weight decay    & None, $\bm{0.01}$, $0.1$ \\
Gradient normalization  & None, $\bm{1.0}$ \\
\bottomrule
\end{tabular}
\caption{Hyperparameter configurations evaluated by grid search. The best configuration on the validation set is highlighted in boldface. 
$_\text{LM}$ indicates the model using conditional label masking.
}
\label{tab:grid_search}
\end{table}
For all models we employ the cased BERT$_{\text{BASE}}$ sentence encoder,
 published by the MDZ Digital Library team (dbmdz)\footnote{ \url{https://huggingface.co/dbmdz/bert-base-german-cased}.}, which has the same architectural setup as the English BERT$_{\text{BASE}}$ counterpart: $12$ multi-head attention layers with $12$ attention heads per layer and $768$-dimensional output embeddings. We initialize all trainable parameters randomly from a normal distribution $\mathcal{N}(0,0.02)$, fix the same random seed of $42$ for all training runs and utilize the AdamW \cite{loshchilov2017decoupled} optimizer with a linear warm-up of $10$\% and a linearly decaying learning rate schedule. Further, we set the width embedding dimension $v$ to $25$, the label embedding dimension $u$ to $128$ (where applicable) and sample a maximum of $N_\text{rel} = 100$ negative relation examples per sentence. In line with Table \ref{tab:grid_search} we also evaluate different levels of dropout regularization before the entity and relation classifier and apply weight decay and gradient normalization. In addition, the models train with varying peak learning rates, batch sizes and prediction thresholds ($\alpha$).
 We train each model variation for 20 epochs and determine its best checkpoint via early stopping\footnote{KPI-BERT's best validation set relation F$_1$-score is achieved in epoch 18.}.


\subsection{Ablation Study}
In the process of hyperparameter selection we paid special attention to certain parameter ablations of KPI-BERT, which are described in Table \ref{tab:modifications}.
\begin{table}[t]
\centering
\scriptsize
\begin{tabular}{lc}
\toprule
Configuration/Ablations & Relation F$_1$ in \%\\
\midrule
KPI-BERT   & $\bm{70.32}$ \\
\hspace{0.5em} $\text{No conditional label masking}$   & $69.25$ \\
\hspace{0.5em} $\text{No filtering overlapping relations}$   & $69.73$ \\
\hspace{0.5em} $\text{No filtering impossible relations}$   & $69.47$ \\
\hspace{0.5em} $\text{No filtering impossible \& overlapping relations}$   & $69.04$\vspace{0.5em} \\
KPI-BERT$_\text{max pooling}$   & $69.16$\\
KPI-BERT$_\text{mean pooling}$   & $69.34$\\

\bottomrule
\end{tabular}

\caption{Ablation study of our tuned  KPI-BERT model, applying different pooling functions and removing filtering heuristics and conditional label masking. 
F$_1$-scores are reported on the validation set since the model ablations are part of a broader grid search.
}
\label{tab:modifications}
\end{table}
First, we find that conditional label masking boosts its validation set relation F$_1$-score by $1.07$ percentage points, which shows the beneficial influence of including prior knowledge in the form of label dependencies in the classification process. Second, we thoroughly investigate the impact of employing different pooling functions. We find that trainable bidirectional GRU-pooling layers outperform the standard mean- and max pooling significantly by $1.16$ percentage points.
Third, we quantify how much our modifications of filtering overlapping and impossible relations improve the model's performance. While both heuristics enhance the relation extraction F$_1$-score, filtering impossible relations leads to a larger improvement, which is expected considering the simplified relation task depicted by the sparsity of Table \ref{tab:relations}.

\subsection{Results}
We retrain the fine-tuned configurations of KPI-BERT and all baselines on the combined training and validation set. To control for a model's susceptibility to random weight initialization, we execute each retraining process 10 times with different seeds.  Thereafter, we evaluate all models on the previously specified hold out test set based on the classification results of the joint named entity recognition and relation extraction task. Table \ref{tab:rel_eval} reports mean and standard deviation of our metrics based on the 10 seed-varying training runs.
\begin{table*}[t]
\centering

\scriptsize
\begin{tabular}{l@{\hspace{1em}}lc@{\hspace{1em}}cc@{\hspace{1em}}cc@{\hspace{1em}}cc@{\hspace{1em}}cc@{\hspace{1em}}cc@{\hspace{1em}}c}
\toprule
in \% &  & \multicolumn{6}{c}{Entity} & \multicolumn{6}{c}{Relation} \\
\cmidrule(lr){3-8}
\cmidrule(lr){9-14}
Name &  Architecture & \multicolumn{2}{c}{Precision$^*$} &  \multicolumn{2}{c}{Recall$^*$} &  \multicolumn{2}{c}{F$_1^*$} &  \multicolumn{2}{c}{Precision} &  \multicolumn{2}{c}{Recall} &  \multicolumn{2}{c}{F$_1$} \\
\midrule
-- & BERT + Linear + RE \cite{taille2020let}      & 76.81 &  (1.00) & 81.57 &  (0.59) & 79.12 &  (0.72) & 66.95 &  (1.51) & 69.34 &  (1.13) & 68.12 &  (1.26)\\
SpERT & BERT + Span + RE \cite{eberts2019span}      & 75.67 &  (0.63) & $\bm{83.45}$ &  ($\bm{0.46}$) & 79.37 &  (0.47) & 67.00 &  (0.84) & 69.48 &  (0.63) & 68.22 &  (0.61)\\
-- & BERT + CRF$_{\text{LM}}$ + RE      & 79.80 &  (0.63) & 82.35 &  (0.51) & 81.05 &  (0.51) & $\bm{70.68}$ &  ($\bm{0.81}$) & 70.62 &  (0.93) & 70.65 &  (0.83) \\
KPI-BERT & BERT + GRU$_{\text{LM}}$ + RE      & $\bm{79.87}$ &  ($\bm{0.55}$) & 82.31 &  (0.55) & $\bm{81.08}$ &  ($\bm{0.53}$) & 70.33 &  (0.55) & $\bm{71.43}$ &  ($\bm{0.60}$) & $\bm{70.88}$ &  ($\bm{0.55}$) \\ 
\bottomrule \addlinespace[1ex]
\multicolumn{14}{l}{\scriptsize{$^*=$ micro average, $_\text{LM}=$ conditional label masking}}
\end{tabular}

\caption{Test set evaluation of the joint named entity and relation classification task, reporting mean (standard deviation) Precision-, Recall- and F$_1$-scores of 10 identical training runs with varying seeds. Our model, KPI-BERT, outperforms the competing state-of-the-art architectures in both entity extraction and relation linking.
}
\label{tab:rel_eval}
\end{table*}
We see that KPI-BERT outperforms the other architectures on both the entity and relation classification objective, yielding respective F$_1$-scores of $81.08$ and $70.88$ percentage points. Noticeably, SpERT and the linearly NER decoding model show a significantly lower mean classification performance on both tasks, which likely originates from neglecting label dependencies when decoding NER tags.
The CRF-based extraction model with conditional label masking (CRF$_\text{LM}$) takes label dependencies into account but still achieves lower scores while suffering from a higher standard deviation across differently seeded runs indicating a worse model robustness compared to KPI-BERT.
\begin{table}[t]
\centering
\renewcommand\tabularxcolumn[1]{m{#1}}
\scriptsize

{\scriptsize
\begin{tabular}{l@{\hspace{1em}}m{6.2cm}r}
\toprule
& Sentence with predicted Entities & Relations \\
\midrule
(a) & Correct relation predictions and annotations. & \\
\midrule
1 & Die {\color{ForestGreen}{[$\underset{\text{kpi}}{\text{sonstigen finanziellen Vermögenswerte}}$}]} enthalten {\color{ForestGreen}{[$\underset{\text{davon}}{\text{Wertberichtigungen}}$}]} in Höhe von {\color{ForestGreen}{[$\underset{\text{davon-cy}}{\text{1,4}}$}]} Mio. \euro{} (Vj. {\color{ForestGreen}{[$\underset{\text{davon-py}}{\text{0,0}}$}]} Mio. \euro{}). & 
\tiny{\color{ForestGreen}{\makecell[r]{kpi -- davon\\ 
davon -- davon-cy \\
davon -- davon-py}}} \\
\midrule
(b) & \multicolumn{2}{@{}m{7.9cm}}{Minor differences between ground truth annotations and model predictions. Arguably, the model predictions are correct, but an annotation mismatch leads to a sentence F$_1$-score below 1.}  \\
\midrule
2 & Der unter Berücksichtigung der Grundsätze des IDW RS HFA 11 {\color{NavyBlue}{[aktivierte}} {\color{Red}{[$\underset{\text{{\color{NavyBlue}{kpi}} kpi}}{\text{immaterielle Vermögensgegenstand}}$}]}{\color{NavyBlue}{]}} in Höhe von TEUR {\color{ForestGreen}{[$\underset{\text{cy}}{\text{4.826}}$}]} wurde beibehalten. & 
\tiny{\color{NavyBlue}{\makecell[r]{kpi -- cy\\
\color{Red}{kpi -- cy}}}} \\
\midrule
(c) & \multicolumn{2}{@{}m{7.9cm}}{Wrong ground truth annotations, but arguably correct model predictions.} \\
\midrule

3 & In den {\color{ForestGreen}{[$\underset{\text{kpi}}{\text{Zinsaufwendungen}}$}]} sind TEUR {\color{Red}{[}}{\color{NavyBlue}{[$\underset{\text{davon-cy \color{Red}{cy}}}{\text{848}}$}]}{\color{Red}{]}} (VJ TEUR {\color{Red}{[}}{\color{NavyBlue}{[$\underset{\text{davon-py \color{Red}{py}}}{\text{816}}$}]}{\color{Red}{]}}) für die {\color{NavyBlue}{[$\underset{\text{davon}}{\text{Aufzinsung von langfristigen Rückstellungen, insbesondere}}$}} {\color{NavyBlue}{für Pensionen, Altersteilzeit und Jubiläum]}} und TEUR {\color{ForestGreen}{[$\underset{\text{davon-cy}}{\text{317}}$}]} (VJ TEUR {\color{ForestGreen}{[$\underset{\text{davon-py}}{\text{432}}$}]}) 
{\color{ForestGreen}{[$\underset{\text{davon}}{\text{Entgelte für Factoring-Geschäfte}}$}]} enthalten.
&
\tiny{\makecell[r]{
\color{NavyBlue}{kpi -- davon} \\ 
\color{NavyBlue}{davon -- davon-cy} \\ 
\color{NavyBlue}{davon -- davon-py} \\ 
\color{Red}{kpi -- cy} \\
\color{Red}{kpi -- py} \\
\color{ForestGreen}{kpi -- davon} \\
\color{ForestGreen}{davon -- davon-cy} \\
\color{ForestGreen}{davon -- davon-py}
}} \\

\bottomrule
\end{tabular}
}
\caption{Several example sentences from the test set with named entity recognition and relation extraction results. {\color{ForestGreen}{Green}}, {\color{NavyBlue}{blue}} and {\color{Red}{red}} represent ``true positive'', ``false positive'', and ``false negative'' entity and relation classifications, respectively. 
}
\label{tab:examples}
\end{table}

Table \ref{tab:examples} showcases several test set sentence examples where KPI-BERT predicts valid relations that either have not been annotated correctly or deviate only slightly from the ground truth entity spans, but still contain valuable information. For instance, in Sentence 2 the annotators did not add the word ``aktivierte'' to the entity annotation of ``immaterielle Vermögensgegenstand''. The model predicted that word, and thus the entity span prediction as well as the relation classifications in this sentence were evaluated as mistakes, although the actual model predictions are arguably correct.  Further, Sentence 3 shows that the model is able to detect long distance relations between entities that even have been wrongly annotated. 

Taking the above findings into account we see that KPI-BERT handles noise in the annotations adequately and is capable of extracting valuable KPI relations from complex sentence structures.


\section{Conclusion and Future Work}

In this paper, we introduce KPI-BERT, an automated system that utilizes new methods of named entity recognition (NER) and relation extraction (RE) to jointly extract and relate key performance indicators (KPIs) and their values from real-world German financial reports. 
Our system leverages a BERT-based architecture that novelly employs a recurrent neural network (RNN) coupled with conditional label masking to sequentially predict KPI tags. In contrast to several other studies, this setup successfully models label dependencies and takes the sequential nature of entity tagging into account. We further integrate a trainable RNN-based pooling layer, which significantly improves upon classic methods like mean and max pooling.

We compare KPI-BERT with three strong relation extraction models, which equally build on BERT embeddings, but differ in their named entity recognition capabilities. Our system outperforms all competing setups in both KPI extraction and entity linking performance, especially surpassing state-of-the-art span-based entity decoders such as SpERT \cite{eberts2019span}.
Ultimately, our results illustrate KPI-BERT's capability to correctly learn and identify long term relations, despite the complexity of the prediction task.

In future work, we plan to evaluate KPI-BERT on public datasets, potentially outside of the financial accounting domain. Additionally, we intend to investigate cross-attention-based transformer architectures coupled with conditional label masking to sequentially tag entities and classify their relations. Further, current state-of-the-art language models like BERT lack numerical reasoning capabilities and are mainly limited to represent plain text. Since we aim to expand the entity and relation extraction task to structured data, e.g. financial tables, a future direction of research will be to replace BERT with a tailored language model, better capable of numerical reasoning and representing tables. 
\section*{Acknowledgment}

This research has been funded by the Federal Ministry of Education and Research of Germany as part of the competence center for machine learning ML2R (01IS18038B/C).




%


\printbibliography

\end{document}